%% file: main_icml2024.tex
\documentclass{article}

\usepackage{microtype}
\usepackage{graphicx}
\usepackage{subfigure}
\usepackage{booktabs} 
\usepackage{comment}
\usepackage{diagbox}
\usepackage{hyperref}

\usepackage[accepted]{icml2024}

\usepackage{amsmath}
\usepackage{amssymb}
\usepackage{mathtools}
\usepackage{amsthm}
\usepackage{makecell}
\usepackage{pifont}
\newcommand{\cmark}{\ding{51}}%
\newcommand{\xmark}{\ding{55}}%

\usepackage[capitalize,noabbrev]{cleveref}

\theoremstyle{plain}

\theoremstyle{definition}

\theoremstyle{remark}

\usepackage[textsize=tiny]{todonotes}

\icmltitlerunning{Position: Leverage Foundational Models for Black-Box Optimization}

\begin{document}

\twocolumn[
\icmltitle{Position: Leverage Foundational Models for Black-Box Optimization}



\icmlsetsymbol{equal}{*}

\begin{icmlauthorlist}
\icmlauthor{Xingyou Song}{gdm}
\icmlauthor{Yingtao Tian}{gdm}
\icmlauthor{Robert Tjarko Lange}{sakana}
\icmlauthor{Chansoo Lee}{gdm}
\icmlauthor{Yujin Tang}{sakana}
\icmlauthor{Yutian Chen}{gdm}
\end{icmlauthorlist}

\icmlaffiliation{gdm}{Google DeepMind}
\icmlaffiliation{sakana}{Sakana AI}

\icmlcorrespondingauthor{Xingyou Song}{xingyousong@google.com}

\icmlkeywords{blackbox, black-box, optimization, foundational, models, llm, language, automl, position, paper, future, challenges, transformer, zeroth, order, derivative, free, BBO, Machine Learning, ICML}

\vskip 0.3in
]



\printAffiliationsAndNotice{}  

\newcommand{\all}[1]{\textcolor{red}{TBD(all): #1}}
\newcommand{\yutian}[1]{\textcolor{blue}{Yutian: #1}}
\newcommand{\yujin}[1]{\textcolor{orange}{\textbf{Yujin: }#1}}
\newcommand{\yingtao}[1]{\textcolor{purple}{\textbf{Yingtao: }#1}}
\newcommand{\robert}[1]{\textcolor{green}{\textbf{Robert: }#1}}
\newcommand{\chansoo}[1]{\textcolor{teal}{Chansoo: #1}}

\begin{abstract}
Undeniably, Large Language Models (LLMs) have stirred an extraordinary wave of innovation in the machine learning research domain, resulting in substantial impact across diverse fields such as reinforcement learning, robotics, and computer vision. Their incorporation has been rapid and transformative, marking a significant paradigm shift in the field of machine learning research. However, the field of experimental design, grounded on black-box optimization, has been much less affected by such a paradigm shift, even though integrating LLMs with optimization presents a unique landscape ripe for exploration. In this position paper, we frame the field of black-box optimization around sequence-based foundation models and organize their relationship with previous literature. We discuss the most promising ways foundational language models can revolutionize optimization, which include harnessing the vast wealth of information encapsulated in free-form text to enrich task comprehension, utilizing highly flexible sequence models such as Transformers to engineer superior optimization strategies, and enhancing performance prediction over previously unseen search spaces.
\end{abstract}


\input{sec1_introduction}

\input{sec2_preliminaries_and_notation}

\input{sec3_related_works}

\input{sec4_challenges_and_techniques}

\input{sec5_future_directions}


\newpage

\section*{Acknowledgements}
We thank Daniel Golovin and Sagi Perel for helpful feedback during the drafting of this manuscript. We further thank Aviral Kumar, Bert Chan, Zi Wang, Frank Hutter, and the AutoML Conference community for early discussions, and Heiga Zen for continuing support. We finally  thank anonymous reviewers for their valuable feedback.

\section*{Impact Statement}
This paper presents work whose goal is to advance the field of 
Machine Learning. There are many potential societal consequences 
of our work, none which we feel must be specifically highlighted here.

\bibliography{refs}
\bibliographystyle{icml2024}


%
%

\end{document}

%% file: sec1_introduction.tex
\section{Introduction}
\label{sec:introduction}

Black-box optimization (BBO) refers to a class of techniques which use minimally observed information to maximize an objective function. Also known as \textit{derivative-free} optimization or \textit{zeroth-order} optimization, the only feedback for an optimizer is the objective value at a given query point, in the absence of additional information such as gradients and second-order derivatives. BBO is broadly prevalent across domains involving experimental design where computing gradients is impossible or infeasible, including automated machine learning~\cite{feurer2015efficient,real2020automl,mellor2021neural}, drug discovery~\cite{turner2021bayesian}, and biological/chemical design~\cite{angermueller2020population}. As an instructive example, in order to improve the performance for a classification task, one may tune a neural network's architecture. Since the accuracy is typically not differentiable with respect to hyperparameters that express design decisions such number of layers, as hyperparameters are varied, one must repeatedly train expensive models over multiple \textit{trials}, to obtain a single accuracy metric. In order to minimize expensive costs, the core challenge in BBO is to \textbf{efficiently} search for parameters which maximize the objective.

Traditional black-box algorithms such as random search~\cite{droste2006upper}, evolutionary methods~\cite{hansen2016cma}, and Bayesian Optimization~\cite{BOreview} have been developed robustly to optimize a wide range of black-box objective functions. Ironically, despite the term ``black-box'', many algorithms inherently perform better based on accurate assumptions about the nature of the objective function. These assumptions, or \textit{priors}, significantly affect the algorithm's key behaviors, such as its predictions over the objective landscape using past observations, and its ability to balance exploration and exploitation to propose the next query point.

\begin{figure*}[t]
    \centering  \includegraphics[width=0.85\textwidth]{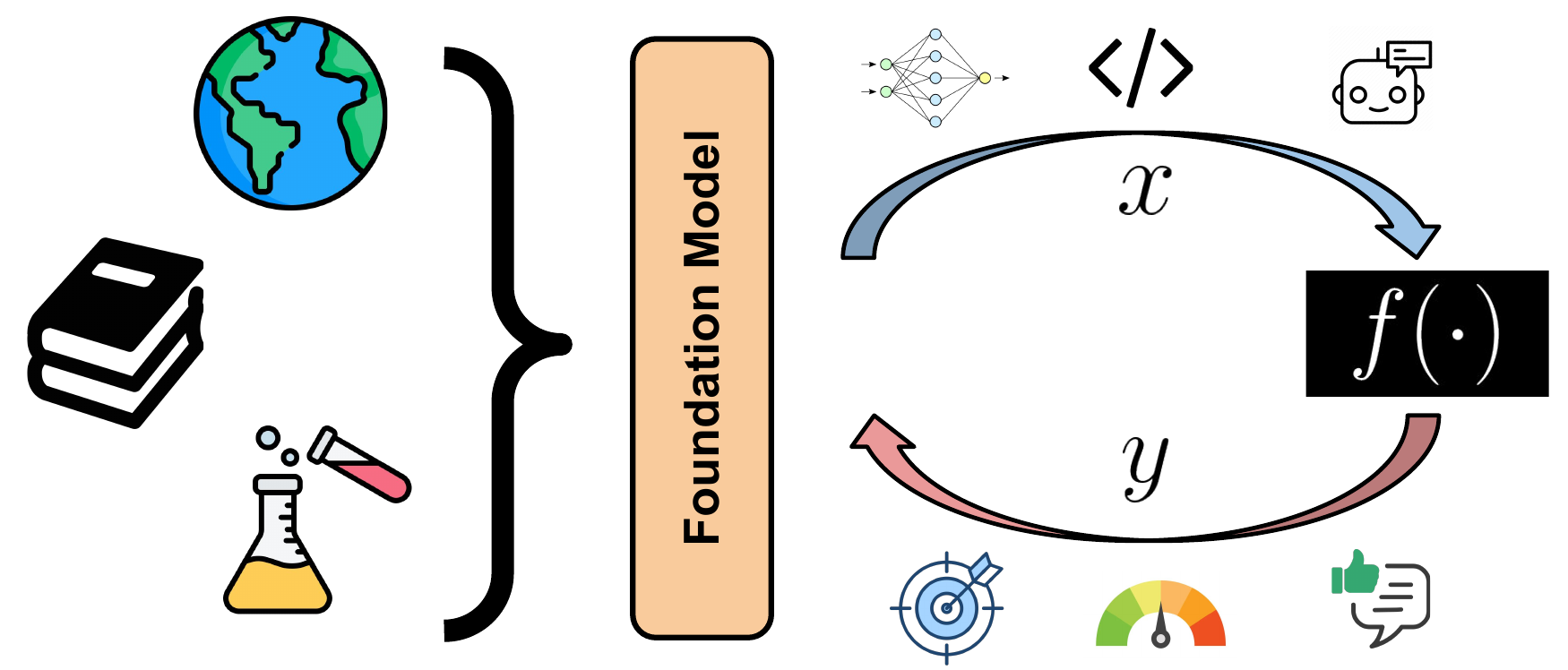}
    \caption{Foundation Models can learn priors from a wide variety of sources, such as world knowledge, domain-specific documents, and actual experimental evaluations. Such models can then perform black-box optimization over various search spaces (e.g. hyperparameters, code, natural language) and feedbacks (numeric values, categorical ratings, and subjective sentiment.}
    \label{fig:front_page}
\end{figure*}

Manually designing these useful priors is difficult and is exacerbated when the prior may not match real world objectives or the prior for one category of tasks may not easily generalize to others, leading to high costs of design. One possible solution is that one may \textit{learn} the prior if provided with realistic objective evaluations, leading to efficient BBO algorithms without needing to explicitly specify the prior. For instance, if given previous evaluations from multiple neural network optimization tasks, one may learn a specific algorithm to better tune the learning rate than a Bayesian optimization algorithm designed more generally for optimizing smooth functions. This type of data-driven approach is known as \textit{learning to optimize}~\cite{li2016learning,chen2017learning,chen2022learning}, a sub-area of meta-learning.

While there have been numerous emergent works based on learning to optimize, the optimization community has not seen wide adoption of these techniques, due to several common obstacles. To list a few:

\begin{enumerate}
    \item \textbf{Lack of generalizability beyond meta-training.} 
    Meta-overfitting~\cite{rajendran2020meta} often prevents learned BBO algorithms from generalizing to test functions, which may have different properties and conditions from ones seen during meta-training, such as smoothness, search space dimensionality, and length of optimization.
    \item \textbf{Low flexibility in exploiting information from data.} BBO is often formulated as a numeric optimization problem, but BBO tasks in practice often contain rich unstructured information beyond numbers, such as task description, search space configuration, historic experiments, textual feedback, and even additional measurements such as learning curves.
    \item \textbf{Little scalability for diverse optimization tasks.} Most learned BBO algorithms are designed for specific types of tasks. A specific algorithm has limited model capacity in handling data from a diverse set of tasks and is typically restricted to a fixed search space. One must train a completely new optimizer for a new optimization problem family, and the fixed search space limitation further limits the training data one could obtain, which exacerbates the scalability issue.
\end{enumerate}

The recent rise of foundation models \citep{foundation_models} via Transformers \cite{transformer} and their use in Large Language Models (LLMs) have stirred an extraordinary wave of innovation in various machine learning domains such as natural language, programming, robotics, and mathematical reasoning. Incorporation has been rapid and transformative, marking a significant paradigm shift where foundation models can be trained on broad data at scale and adapted to a wide range of tasks. In contrast, such methods have not been thoroughly studied in the area of BBO, even though applicable training data exist, consisting of not only optimization trajectories but also relevant world knowledge on experimental design and optimization. Such knowledge could adapted to BBO tasks over various search spaces and data types, as illustrated in Figure \ref{fig:front_page}.

\textbf{This position paper advocates for wider research and adoption of Transformers and LLMs for black-box optimization}, as they possess a few key benefits which are desirable to address the aforementioned challenges in learned BBO:

\begin{enumerate}
    
\item The Transformer's input format allows \textbf{modeling a wide variety of data}, ranging from fixed dimensional vectors to sequences of text and even multi-modal data, given the proper encoding schemes. This input format allows the possibility of learning a single model from mixed datasets of diverse optimization tasks, each with task-specific side information. 
\item The \textbf{Transformer's superior scalability} is a much needed property for learning a general BBO algorithm from large datasets compared to traditional machine learning models. 
\item The Transformer's \textbf{in-context learning capacity} is useful for improving a learned algorithm's generalizability to new settings.
\end{enumerate}



Our paper is structured as follows: Section~\ref{sec:preliminaries_notation} provides preliminaries and notation on BBO, and Section~\ref{sec:previous_works} conducts thorough survey of previous works, organized by approaches with gradually increasing generalities and relationship with sequence-based models, ultimately towards LLM-based techniques. In Section~\ref{sec:challenges_techniques}, we then identify challenges and advocate for techniques which are crucial in advancing the field of learned optimizers. These include data-driven training, better data representations and multi-modality, and more flexible model prompting, with additional emphasis on improved BBO benchmarking. Lastly, in Section~\ref{sec:future_directions}, we provide a vision for next generation LLM-based optimizers, which are able to integrate multi-modal data, incorporate user-provided feedback, and manage problem-specific information over long contexts.

%% file: sec2_preliminaries_and_notation.tex
\section{Preliminaries and Notation}
\label{sec:preliminaries_notation}
\newcommand{\desc}{m}
\newcommand{\DESC}{\mathcal{M}}
\newcommand{\dimension}{d}

\begin{figure*}[t]
    \centering
    \includegraphics[width=0.95\textwidth]{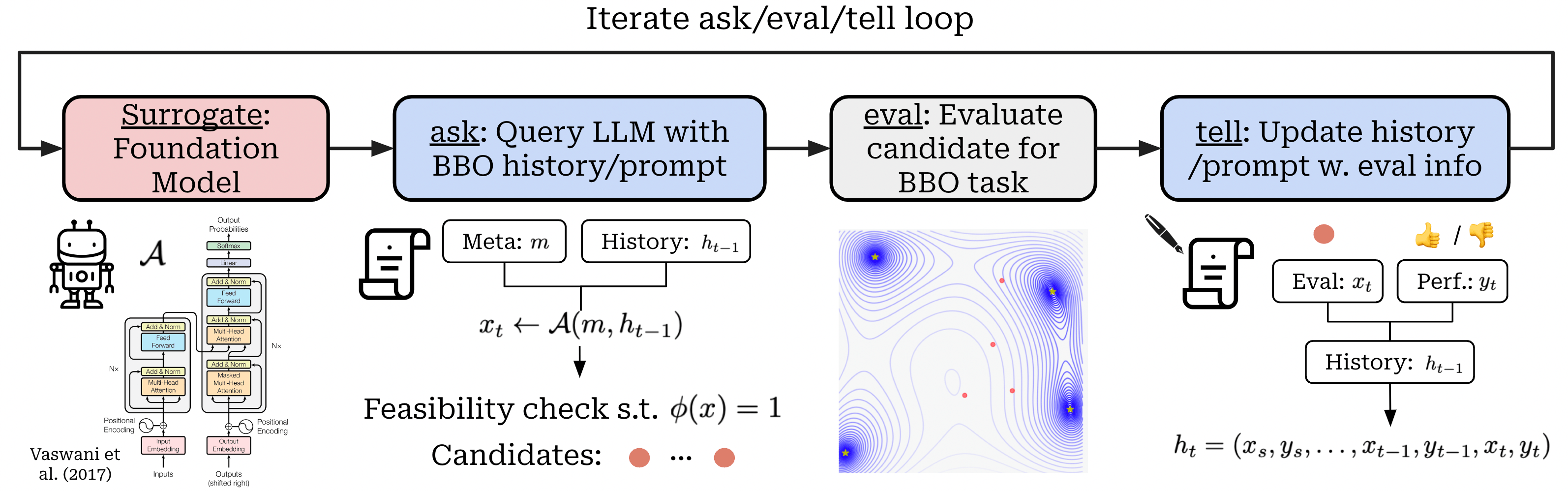}
    \caption{Black-box optimization loop with sequential foundation models. Using metadata $m$ and history $h$, the model proposes candidates $x$ which are checked for feasibility, evaluated, and then appended to the history.
    }
    \label{fig:bbo_loop}
\end{figure*}


\begin{algorithm}[h]
    \floatname{algorithm}{Problem}
    \caption{Experimental Design Problem}
    \begin{algorithmic}
     \REQUIRE Problem meta-description $\desc \in \DESC$, search space $\mathcal{X}$, and feedback function $f \in \mathcal{F}$, Algorithm $\mathcal{A}$
     \STATE Initialize $\mathcal{A}$ using all provided information.
     \STATE Initialize history $h \gets []$
     \FOR{t=1, 2, \ldots \text{ {\bf until} end condition is met}}
       \STATE Generate a suggestion: $x_t \gets \mathcal{A}(\desc, h_{1:t-1})$
       \STATE Receive feedback: $y_t \gets f(x_t)$
       \STATE Update history: $h \gets$ concatenate($h, x_t, y_t$).
     \ENDFOR
    \STATE \textbf{Return $h_{1:t}$ as $H$}
    \end{algorithmic}
\end{algorithm}
\raggedbottom

We begin by defining the generic problem of experimental design. At the $t$-th iteration, the algorithm $\mathcal{A}$ suggests $x_t$ in the search space $\mathcal{X}$ and receives feedback $y_t = f(x_t)$, from a feedback space $\mathcal{Y}$. We use $h_{s:t}$ to denote the optimization \emph{history} from the $s$-th iteration up to the $t$-th iteration, i.e. the sequence of suggestions and feedbacks $(x_s, y_s, x_{s+1}, y_{s+1}, \ldots, x_t, y_t)$. We may omit the subscript for $h$ if it is clear from the context. We use $\mathcal{H}$ to denote the space of all possible trajectories of any length, and uppercase $H$ without the subscript to denote the full trajectory of the run.

Note that what makes our problem formulation different from the standard black-box optimization problem is the existence \emph{problem meta-description}. This meta-description $m$ contains some information that may hint towards a more specific subset of the function set $\mathcal{F}$ which $f$ was chosen from. 


In this paper, we often consider the setting where we have access to a database of historic runs, where each element is a tuple $(\desc_i, \mathcal{X}_i, H_i, \mathcal{A}_i)$; note that the feedback function $f_i$ itself is \emph{not} included.  




\subsection{Search Spaces as Sequences}
\label{subsec:search_space_sequences}
To better understand our motivation for using foundation models, one important insight is to connect search spaces to the notion of \textit{formal grammars} \cite{automata_theory}. Without introducing overly technical background on grammars, it suffices to see that $\mathcal{X}$ is representable as a sequence of atomic \textit{parameter configurations}, or alphabets $\mathcal{X}^{(1)}, \mathcal{X}^{(2)}, \ldots$ and thus each $x$ can be seen as a collection of parameters, or a ``string" $(x^{(1)}, x^{(2)}, \ldots)$. In addition, there may be a \textit{feasibility function} $\phi: \mathcal{X}^{(1)} \times \mathcal{X}^{(2)} \times \ldots \rightarrow \{0, 1\}$ which determines whether $x$ is admissible in the search space.

Note that while a search space $\mathcal{X}$ can be commonly reparameterized into a new space $\mathcal{X}'$, certain fundamental invariants remain which dictate the inherent properties of the space. These are:

\begin{itemize}
\setlength{\parskip}{0pt}
\setlength{\itemsep}{3pt}
\item \textit{Parameter Type:} $\mathcal{X}^{(i)}$ can be continuous (e.g. subinterval of $\mathbb{R}$) or discrete (e.g. $\mathcal{X}^{(i)} \subseteq \{1, 2, \ldots, n\}$).
\item \textit{Constraints:} $\mathcal{X}$ can be unconstrained or may possess a nontrivial $\phi$. 
\item \textit{Length Boundedness:} $x$ may be of unbounded length $(x^{(1)}, x^{(2)}, \ldots) $ or explicitly bounded (both below and above).
\end{itemize}

The particularly nuanced invariant is $\phi$, which is important to organize into the following categories:

\begin{itemize}
\setlength{\parskip}{0pt}
\setlength{\itemsep}{3pt}
\item \textbf{Unconstrained:} $\phi(x) = 1$ always.
\item \textbf{Inductive:} $\phi$ can be factored locally as $ \prod_{j=0} \phi (x^{(j+1)} | x^{(1:j)}) $ where $\phi(x^{(j+1)} | x^{(1:j)})$ is efficiently computable and can be seen as a one-step restrictor mapping $x^{(1:j)} = (x^{(1)}, \ldots x^{(j)})$ to a subset of allowed values in $\mathcal{X}^{(j+1)}$.
\item \textbf{Multi-step Inductive:} The above factorization requires use of $k$-step restrictors $\phi(x^{(j+1:j+k)} | x^{(1:j)})$ where $k\ge 2$ is reasonably small.
\item \textbf{Global:} If $\phi$ does not admit an efficiently computable factorization, then determining feasibility only occurs once the entire $x$ is formed, and $\phi$ can be seen as a global ``compilation check". 
\end{itemize}

%% file: sec3_related_works.tex
\section{Previous Works and Motivation} 
\label{sec:previous_works}

\subsection{Previous Applications}
While it is natural to organize optimization problems by application domain, it is far more useful for researchers to organize problems by their fundamental search spaces and invariants defined in Section \ref{subsec:search_space_sequences}, which heavily influence algorithm design. We organize these problems in Table \ref{table:applications_search_space}.

\begin{table}[h]
\centering
\renewcommand{\arraystretch}{1.6}
\scalebox{0.88}{
\begin{tabular}{|l|c|c|c|}\hline
\diagbox[height=3\line,width=20ex]{Application}{Invariant} & Parameters & Constraints & \makecell{Length\\Bound} \\
\hline
Traditional & Any & Any & Bounded  \\
\hline
Combinatorial & Discrete & Inductive & Bounded \\
\hline
\makecell[l]{Genetic\\Programming} & Any & \makecell{Multi-step\\Inductive} & Any \\
\hline
Free-form Code & Discrete & Global & Unbounded \\
\hline
Free-form Prompt & Discrete & None & Unbounded \\
\hline
\end{tabular}
}
\caption{Example applications categorized by search space invariants.}
\label{table:applications_search_space}
\end{table}

By organizing using $\phi$, we can see that traditional BBO problems such as continuous optimization \cite{elhara2019coco} and protein sequence design \cite{angermueller2020population} have consisted of unconstrained Cartesian spaces. In contrast, combinatorial problems such as Traveling Salesman \cite{flood1956traveling} and other graph problems \cite{balakrishnan2012textbook} use primitives such as permutations and combinations, which are inductive constraints, as the next possible values for $x^{(j+1)}$ are determined by previously chosen $x^{(1:j)} \subset \{1, \ldots, n\}$.

In the domain of program search, classic genetic programming \cite{real2020automl} factorizes the search space by smaller subtrees of possible symbols, and thus use multi-step inductive constraints. However, free-form code search \cite{romera2023mathematical} is constrained by both compilation and runtime errors, which can only be checked after an entire program is globally created. Ignoring optional grammar constraints, ultimately the field of prompt optimization defines objectives over arbitrary strings. 

\definecolor{darkgreen}{rgb}{0.3, 0.8, 0.4}
\definecolor{darkred}{rgb}{0.8, 0.2, 0.2}
\newcommand{\greencheck}{\textcolor{darkgreen}{\cmark}}
\newcommand{\redx}{\textcolor{darkred}{\xmark}}

\begin{table*}[t]
\centering
\renewcommand{\arraystretch}{1.5}
\scalebox{0.95}{
\begin{tabular}{|c|c|c|c|c|c|c|c|c|}\hline
\diagbox{Method}{Capabilities} & \makecell{In-task \\ Adaptability} & \makecell{Across-Task \\ Transferrable} & \makecell{In-context \\ Learning} & \makecell{Search Space \\ Transferrable} & \makecell{Length \\ Scalable} & \makecell{Multi-\\modality} & \makecell{Built-in \\ NLP} \\
\hline
Non-learnable & \redx & \redx & \redx & \redx & \redx & \redx & \redx \\
\hline
Feature-based & \greencheck & \redx & \redx & \redx & \redx & \redx & \redx \\
\hline
Meta-learned & \greencheck & \greencheck & \redx & \redx & \redx & \redx & \redx \\
\hline
Sequential History & \greencheck & \greencheck & \greencheck & \redx & \redx & \redx & \redx \\
\hline
Sequential Search Space & \greencheck & \greencheck & \greencheck & \greencheck & \redx & \redx & \redx \\
\hline
Attention-based & \greencheck & \greencheck & \greencheck & \greencheck & \greencheck & \redx & \redx \\
\hline
Token-based & \greencheck & \greencheck & \greencheck & \greencheck & \greencheck & \greencheck  & \redx \\
\hline
LLMs & \greencheck & \greencheck & \greencheck & \greencheck & \greencheck & \greencheck & \greencheck \\
\hline
\end{tabular}
}
\caption{Classes of methods organized by their capabilities. Note: Method names based on increasing development order - e.g. ``Attention-based" can consist of techniques up to their development such as meta-learning, but not LLMs.}
\label{table:method_capabilities}
\end{table*}

\subsection{Previous Techniques}
From our sequential formalization of both the experimental design problem and their underlying search spaces, it is thus natural to see the relevance of models which use sequential representations, particularly for more complex applications. We outline key previous works in order of development, which have crucially developed along this direction, with their capabilities shown in Table \ref{table:method_capabilities}, eventually leading towards the use of learned foundation models in BBO loops, as illustrated in Figure \ref{fig:bbo_loop}.

\textbf{Non-learnable:} These consist of purely hand-designed rules for proposing $x_{t+1}$ based on $h_{1:t}$ and can be deterministic or stochastic. A common theme in the evolutionary algorithm literature is around the idea of \textit{mutation}, in which only a fixed-size pool of historical trials are used to construct $x_{t+1}$ by changing every parameter in $(x_{t+1}^{(1)}, x_{t+1}^{(2)}, ...)$ using perturbations and crossover. Such methods are usually sample inefficient, as many algorithmic components are based on unguided randomness.

\textbf{Feature-based:} Traditional model-driven BBO requires formatting $(x,y)$ as \textit{features}, in particular creating representation mappings $\mathcal{X} \rightarrow \mathbb{R}^{d}$ to allow the construction of learnable regressors $\mathbb{R}^{d} \rightarrow \mathbb{R}$ such as Tree-structured Parzen Estimator (TPE) \citep{Bergstra2011TPE}, Random Forests \cite{hutter2011sequential}, Gaussian Processes (GPs) \citep{snoek2012practical}, and feed-forward neural networks to guide search. While feature construction is straightforward for Cartesian spaces, it is considerably more complex for conditional spaces such as combinatorics \cite{combo} or graph spaces \cite{nasbowl}. 

Generally, feature-based methods have allowed the algorithm to adapt to incoming observations within the current task, although their performance is bounded by the limited number of available observations and their hand-designed prior.

\textbf{Meta-learned:} In order to take advantage of information from observations of related tasks, transfer-learning / meta-learning algorithms are proposed to learn task-independent prior and optional task-specific parameters. Representative feature-based meta-learned methods often combine Gaussian Processes with deep-learning kernels \citep{swersky2013multi,yogatama2014efficient,perrone2018scalable,volpp2019meta,wistuba2021few,wang2021pre}. While more specific prior can be learned in this approach, in-task finetuning is still required for optimal performance \cite{wistuba2021few}.

\textbf{Sequential History:} In order to utilize a static model which can explicitly process $H$ and directly output a $x$, a sequential model such as a recurrent neural network (RNN) introduces a weight-indepdendent sequence axis for additional \textit{in-context learning}. For the multi-turn bandit case, \cite{chen2017learning} places the historical trajectory $H$ on the sequence axis to propose a new $x$. Thus the model's weights are independent of the length of $H$ and can arbitrarily process any history.

%

\textbf{Sequential Search Space:} Sequential models are also used to process search spaces with an inductive $\phi$. \cite{rl_co} uses an RNN's length axis along with masking based on $\phi$, to construct a distribution $p_{\theta}(x) \in \mathcal{P}(\mathcal{X})$ over a family of combinatorial spaces. This can be seen as a form of restricted decoding over the parameters $(x^{(1)}, x^{(2)}, ...)$ of $x$. By placing the parameters $x$ along the length axis, the model's weights are thus independent of the size of $x$ and the search space.

\textbf{Attention-based:} Since RNNs have difficulty scaling to accommodate long sequences, the attention mechanism has been adopted by e.g. \citet{lange2023discovering,lange2023discovering2} to construct evolutionary and genetic algorithms possessing strong generalization capabilities. 
In addition, \citep{muller2023pfns,muller2021transformers, transformer_neural_processes} train Transformers to conduct in-context Bayesian inference as a surrogate model for Bayesian optimization, while \citep{expt, gpt_bbo} use Transformers to directly propose suggestions $x$.
    
\textbf{Token-based:} In order to obtain weight-independence for both the history \textit{and} search space, the OptFormer \cite{chen2022towards} represents every parameter value as a \textit{token} to avoid constructing features altogether. Metadata was additionally tokenized and processed as text. 
Despite the use of text as input, the model was trained on a relatively small text corpus and did not demonstrate the emergent capacity to understand semantic meanings of metadata yet. \cite{lange2024evolution} alternatively learns its own embedding to map each $x^{(i)}$ to a single vector.

While there have yet to be optimizers which process other modes of data such as images and audio, token-based methods allow the use of unified vocabularies which can represent multiple modes simultaneously, which may be of interest in the future for representing more exotic forms of $m,x,y$.





%

\textbf{Natural Language:} LLMs provide a powerful general-purpose tool for leveraging additional text-based BBO information. There exist several applications using off-the-shelf pre-trained models as mutation operators in the context of code-based settings.
More specifically, \cite{lehman2023evolution} embedded LLMs into Genetic Programming, i.e. the model serves as a mutation operator optimizing morphologies and behaviors on the code level. 
\cite{chen2023evoprompting, llmatic} use an LLM to adaptively mutate and cross-over code for evolutionary Neural Architecture Search.
Furthermore, \cite{meyerson2023language} introduce Language Model Crossover where they concatenate parent solutions into a prompt and collect offsprings to evaluate from the text-based output. Here, the LLM has to act as a variation operator and thereby evolves genomes representable as text strings \cite{liu2023algorithm}.

While these approaches were mostly applied to fairly narrow use-cases, more recent works have been leveraging LLMs as general-purpose optimizers. \cite{yang2023large} first leveraged LLMs as pure black-box optimizers. More specifically, they showed that PaLM-2 \citep{anil2023palm} models to directly output candidates and perform linear regression, prompt optimization, and other tasks. 
LLMs can also be used to propose distributions for sampling $x$'s, as shown for single-objective \citep{lange2024large, liu2023large} and multi-objective \cite{liu2023large_multi} evolutionary optimization.
\cite{zhang2023using, nie2023importance} further showed that LLMs can be applied to hyperparameter optimization problems using natural language instructions and feedback. 
Additionally, LLMs can be used for related subtasks such as multi-task regression \citep{omnipred} and surrogate modeling \citep{llm_bayesian}.



%% file: sec4_challenges_and_techniques.tex
\section{Challenges and Required Techniques}
\label{sec:challenges_techniques}
\begin{figure*}[h]
    \centering
    \includegraphics[width=0.85\textwidth]{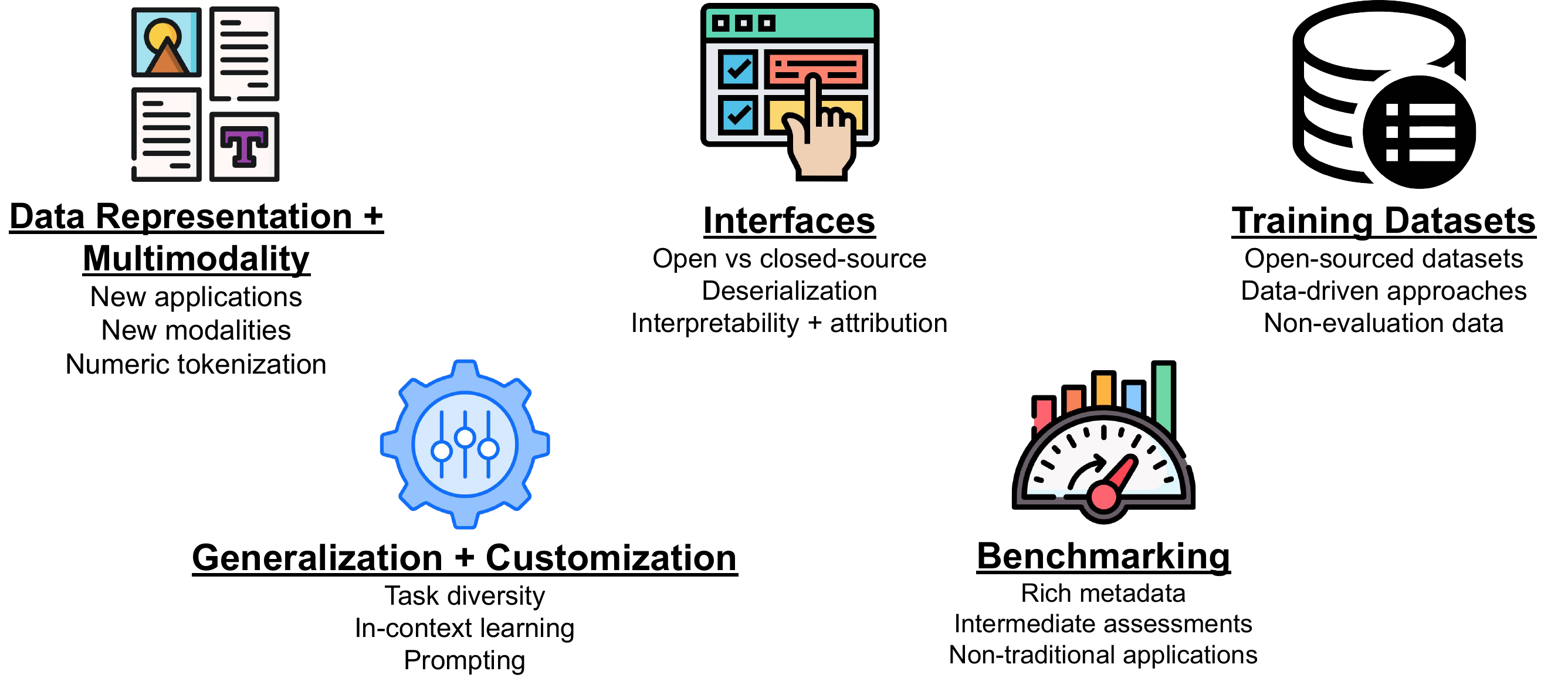}
    \caption{Summary of future challenges and open questions for BBO with LLMs.}
    \label{fig:overview}
\end{figure*}


Despite the significant potential benefits of learned optimizers, so far they have not been widely used, let alone adopted in production-grade optimization systems \cite{google_vizier, ax}. We believe this is due to a variety of key technical challenges, many of which can be resolved under the current status-quo of using Transformer-based foundation models, while other challenges will require further improvements, summarized in Figure \ref{fig:overview}.

\textbf{Data Representation and Multimodality:} Most meta-learning based BBO methods consider a fixed search space $\mathcal{X}$ and feedback space $\mathcal{Y}$ \citep{swersky2013multi,wistuba2021few,chen2017learning}.
Specific embedding methods must be manually crafted for each domain (hyperparameters, graphs, code, etc.) This drastically limits the scope in which a learned method can be transferred over. As a result, an optimizer must be learned from scratch for every new type of application or even search-space dimension.

In contrast, sequence-based and more broadly text-based representations, when consumed by Transformer models and applied over both the historical and search space axes of optimization problems, greatly increase generality and transferrability. More broadly is the ability to incorporate multimodal knowledge, outside of using textual representations. For instance, one may be able to predict the outcome of a machine learning experiment not only from observing hyperparameters, but also the code used for training, neural network parameters \citep{nn_weight_prediction}, and even the dataset used. Another example in a wet-lab scenario could be to express $(m,x,y)$ as images, which can be more expressive than pure text or numbers.
While modality-specific encoders for BBO have yet to be applied, general usage over other domains has been widely studied and achieved, such as across-domain transfer \citep{raffel2020exploring,brown2020language} and across-modality alignment \citep{radford2021learning,girdhar2023imagebind}.

Open-questions remain on the optimal tokenization of optimization-specific domains, especially for numeric and mathematical objects. Currently, general consensus from previous literature \citep{optformer} suggests that numeric objects (e.g. floats, graphs) should at least be tokenized according to their unit building blocks (e.g. digits, vertices). Thus it is not clear that representing numbers in standard human-readable format is optimal, especially as certain tokenizers \citep{kudo2018sentencepiece} will not split such numbers digit-by-digit --- for example \texttt{123.4} may be split into tokens representing \texttt{[12, 3., 4]}. 

\textbf{Interfaces:} One would expect that due to common fundamental ideas behind LLMs, their user interface should also remain consistent. This is not true in practice however, as multiple different LLMs provide APIs with varying degrees of interactivity. For instance, open-source LLMs \citep{llama2} allow the user to fine-tune the model against custom data, while closed-source LLMs \citep{gpt4} only provide remote procedure calls. Furthermore, many closed-source services utilize different technical choices when defining an ``embedding" or when performing inference and decoding. While there has been some recent effort for creating unified APIs \cite{cheng2023batch,Pham_OpenLLM_Operating_LLMs_2023}, these have mostly been for the purpose of prompting. 

For optimization in particular, currently there is no unifying format which the community agrees upon. This is further constrained by a strong need to \textit{deserialize} string outputs back into $x$ and $y$. While LLMs sufficiently trained on code data can output tabular JSON formats, e.g. \texttt{\{batch\_size:128,learning\_rate:0.1\}} used in hyperparameter optimization, following more sophisticated combinatorial and numeric constraints is still an open question.

There is additional interest in functionality beyond regular $x$-proposal or $y$-regression for BBO. While the LLM community has actively studied interpretability \citep{llm_explainability}, attribution \citep{llm_attribution}, and uncertainty quantification \cite{llm_uncertainty} for general LLM assistant scenarios, there has been little work in the context of BBO and decision-making in bandit settings. Such BBO-specific functionalities are likely to be more difficult to induce, as they require advanced mathematical and numerical reasoning beyond current LLM capabilities.


\textbf{Training Datasets:} Learned BBO algorithms are significantly dependent on both the quality and quantity of data available. The most obvious useful data are function evaluations $(x,y)$, as they can be used to form a prior on the nature of the function being optimized. The most common form of usage are for pretraining regressors which can then guide the search process. Due to limitations to fixed search spaces, traditional methods do not have or consider large real-world training datasets. As a result, there is currently a lack of large-scale open-source evaluation datasets. This consequently limits a learned BBO method's generality, even though \citet{chen2022towards} is the first work which has shown that training a foundation model over large-scale hyperparameter tuning data collected by Google Vizier~\cite{google_vizier} leads to robust generalization to new tuning tasks with different search and feedback spaces. 

Unfortunately, industry datasets such as Google Vizier \cite{google_vizier} and Ax \citep{ax} with rich task-specific metadata, are unlikely to be fully open-sourced due to proprietary legal and privacy concerns. While there have been some efforts to standardize smaller-sized public datasets \cite{eggensperger2021hpobench, trabucco2022design} and centralize community-driven data over an open platform such as OpenML \cite{OpenML2021}, an important question is when the BBO community may be able to truly embrace data-driven approaches.

A further open-question remains on how to obtain more domain knowledge besides using function evaluations. For instance, techniques from protein design \citep{protein_generation_transformer, llm_generate_protein} train foundation models over protein sequence data as a form of knowledge absorption. It remains to be seen whether additional ``optimization knowledge" can be gained from free-form datasets such as textbooks, academic papers, and whether these can lead to emergent behaviors via generative pretraining and RL-finetuning. For example, the field of hyperparameter optimization could be improved by semantically understanding the nature of the tuned objective - e.g. academic knowledge of the term ``batch size" could lead to better hyperparameter proposals. 

\textbf{Generalization and Customization:} The primary challenge of meta-learning is its limited generalization to new tasks from different domains, search space dimensions, constraints, or simply different trajectory lengths. This is mostly due to the issue of \textit{meta-overfitting} where the optimizer has overfitted to a limited number of optimization tasks with low diversity. Expanding the training set of tasks and optimizer model size could be effective solutions as suggested by LLM scaling laws. Additionally, the impressive in-context learning capacity \citep{brown2020language} of a large Transformer allows a pretrained model to generalize to a new setting with few demonstrations without changing its weights. This would be an intriguing feature for adapting meta-learned BBO methods to new application tasks as it is typically difficult to obtain a large number of similar optimization tasks for adaptation in real applications.

Related to generalization is the customization of a pretrained optimizer to different use cases. It is necessary that a model understands the user's \textit{intent} to optimize a function. For example, a user may prefer more exploration in a preliminary experimental design phase with a small budget while another user prefers to find a good solution as soon as possible. Furthermore, different users may require different safety / exploration constraints and provide various side information about the objective. Training a new optimizer for every use case is not feasible in practice. In contrast, the sequential input format of LLMs permits efficiently customizing its behavior through prompting, which can provide additional capabilities such as providing explanations for why the model suggested $x$ or confidence over predictions of $y$-values.

While previous works \cite{chen2021decision,chen2022towards,lange2024evolution,lange2024large} have used Transformers and even LLMs as the underlying model, they still only \textit{implicitly} learn the intent for improvement by manipulating the sequential order of trials, such as least-to-most sorting by $y$-values and inversely using a query $y$ to obtain a response $x$. The next generation of optimizers must allow \textit{explicit} commands for optimization-related behaviors.


\textbf{Benchmarking:} Currently, most of BBO benchmarking requires well-defined, simple search spaces and objectives with minimal contexts, to follow the namesake ``blackbox" and provide fair comparisons between general-purpose methods. These include extensive and official benchmarking tasks such as BBOB~\cite{elhara2019coco}, NASBench~\cite{nasbench101} or NeuroEvoBench~\cite{lange2023neuroevobench}. However, we advocate for a wider spectrum of benchmarks that do not follow this regime, as they are crucial to the future of BBO. 

For starters, there is a need for \textbf{metadata-rich black-box optimization}, especially as real-world problems inherently are not truly black-box. These benchmarks should be designed to emphasize an optimizer's utilization of rich metadata. In particular, this alludes to the use of additional language encoders in optimizers not only limited to LLMs, such as ~\cite{lange2024large, chen2022towards} which have shown promise over in real-world problems by using dedicated decoders. 

As systems become more integrated with LLMs end-to-end, there is also a need to assess intermediate decision-making capabilities which are not typically measured by opaque traditional BBO metrics such as regret or best-objective. As shown recently via multi-armed bandit problems, \citep{llm_mab} claims LLMs such as GPT-4 to be quite poor at balancing explore-exploit tradeoffs. Meanwhile, \citep{omnipred} demonstrated strong results on regression for fine-tuned language models. Another important capability is constraint-following, not just over mathematical objects but also ambiguous natural language constraints (e.g. ``Give a batch size which fits into GPU memory").

Furthermore, there are important yet non-obvious applications that can be framed as black-box optimization, particularly over more exotic search spaces or feedback formats. For example, there has been recent interest in evaluating and benchmarking LLM-based code generation~\cite{ji2023benchmarking,fu2023mme,xu2022systematic}. Human-based reward modeling and reinforcement learning from human feedback (RLHF) can also respectively be seen as regression and black-box surrogate optimization~\cite{ramamurthy2022reinforcement,wang2018glue}. In addition, heavily-used techniques such as prompt engineering possess domain-specific guidelines~\cite{wang2023prompt,WANG2023100047} but still lack well-accepted benchmarks and evaluation paradigms. Ultimately it remains an open problem to integrate these applications into a setting comparable to that of traditional BBO problems.






%% file: sec5_future_directions.tex
\section{Future Directions}
\label{sec:future_directions}


We identified and discussed challenges hindering progress in applying LLM and Transformer-based models specifically over optimization tasks, as well as their existing solutions in Section~\ref{sec:challenges_techniques}. These insights highlight key areas where focused research efforts can lead to substantial advancements. Venturing further into a domain that permits a more extended time horizon and a wider scope reveals intriguing and promising directions for future research.

The envisioning of a universal LLM, adept in both natural language understanding and executing complex optimization tasks, marks a significant leap forward in AI technology. Such a forward-looking model, with its capacity to \textbf{process user-provided metadata about target optimization problems} through straightforward instructions, promises transformative impact across numerous sectors. Imagine its application in enhancing human-robot interaction~\cite{brohan2023rt,yu2023language}, revolutionizing autonomous driving systems~\cite{sha2023languagempc}, innovating reward design in learning environments~\cite{ma2023eureka}, and redefining efficiency in logistics planning~\cite{li2023large}. Moreover, the scope of this model's capabilities extends well beyond these areas. Envision, for instance, its application in a multi-agent setting where, through self-dialogue, the model could autonomously uncover and propose solutions to complex problems that elude human detection. Such innovative capabilities hint at a promising trajectory towards the realization of artificial general intelligence.

Realizing the vision of a universal LLM for complex optimization tasks hinges on overcoming significant hurdles, but recent promising strides in related areas signal a path forward. One of the primary challenges is \textbf{managing long context lengths in problem descriptions}, a task where methods like those in ~\cite{jin2024llm,xiao2023streamingllm} show promise, yet their direct application to optimization-focused LLMs remains to be validated. Equally important is the \textbf{integration of multi-modal data}, crucial for a holistic understanding of complex problems, with current techniques in~\cite{liu2023llava,liu2023improved} potentially serving as a foundation. Furthermore, the idea of model composition, perhaps involving a network of specialized LLMs as indicated by research in~\cite{shen2023mixture,bansal2024llm}, opens new avenues but also demands further exploration. Crucially, these challenges underscore the need for enhanced collaboration and open research, urging a collective approach across different LLMs and disciplines to unlock the full potential of such models.